\documentclass[journal]{IEEEtran}
\usepackage{times}
\usepackage{epsfig}
\usepackage{graphicx}
\usepackage{amsmath}
\usepackage{amssymb}
\usepackage{booktabs}
\usepackage{subfigure}
\usepackage{blindtext}
\usepackage{caption}
\usepackage{xspace}
\usepackage{xcolor}

\makeatletter
\DeclareRobustCommand\onedot{\futurelet\@let@token\@onedot}
\def\@onedot{\ifx\@let@token.\else.\null\fi\xspace}

\def\eg{\emph{e.g}\onedot} 
\def\ie{\emph{i.e}\onedot} 
 
\def\etc{\emph{etc}\onedot}

\makeatother

\ifCLASSINFOpdf
\else
\fi
\begin{document}
%
\title{\emph{RXFOOD}: Plug-in RGB-X Fusion for Object of Interest Detection}
%
%
%

\author{Jin Ma$^{\dagger}$,
        Jinlong Li$^{\dagger}$, 
        Qing Guo,
        Tianyun Zhang,
        Yuewei Lin$^{*}$,
        Hongkai Yu$^{*}$ 
\thanks{Jin Ma, Jinlong Li, Tianyun Zhang and Hongkai Yu are with Cleveland State University, Cleveland, OH, 44115, USA. Qing Guo is with the Centre for Frontier AI Research (CFAR) and Agency for Science, Technology and Research (A*STAR), Singapore, and the Institute of High Performance Computing (IHPC), Agency for Science, Technology and Research (A*STAR), Singapore. Yuewei Lin is with Brookhaven National Laboratory, Upton, NY, 11973, USA.}
\thanks{$\dagger$ indicates the co-first authors. * Corresponding authors: Yuewei Lin (e-mail: ywlin@bnl.gov) and Hongkai Yu (e-mail: h.yu19@csuohio.edu).}
}
\maketitle

\begin{figure*}[t]
    \centering
    \includegraphics[width=1\textwidth]{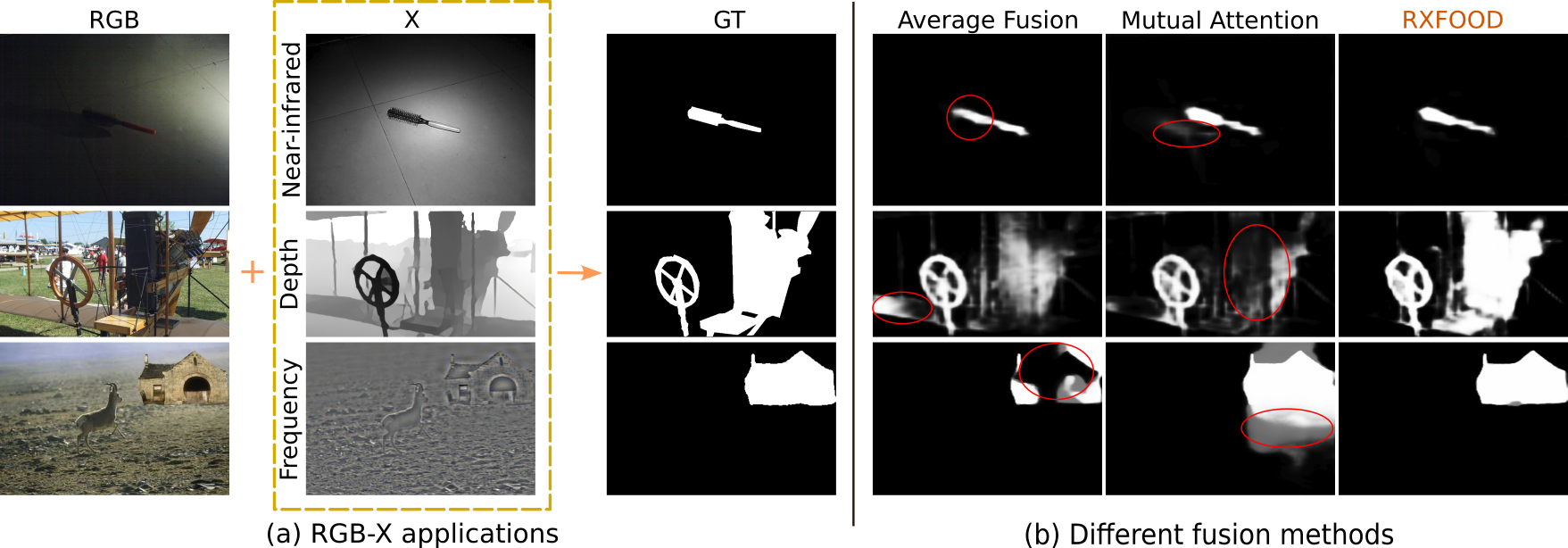}
    \caption{RGB-X Object of Interest Detection: (a) Diverse multimedia applications, from top to bottom: RGB-NIR salient object detection, RGB-D salient object detection, RGB-Frequency image manipulation detection, whose X inputs are Near-infrared, Depth, and Frequency component respectively. (b) Detection results by different fusion methods: Average fusion, Selective Self-Mutual Attention fusion~\cite{liu2020S2MA}, and the proposed RXFOOD.} 
    \label{fig:motivation} 
\end{figure*}

\begin{abstract}
   The emergence of different sensors (Near-Infrared, Depth, \etc) is a remedy for the limited application scenarios of traditional RGB camera. The RGB-X tasks, which rely on RGB input and another type of data input to resolve specific problems, have become a popular research topic in multimedia. A crucial part in two-branch RGB-X deep neural networks is how to fuse information across modalities. Given the tremendous information inside RGB-X networks, previous works typically apply naive fusion (\eg, average or max fusion) or only focus on the feature fusion at the same scale(s). While in this paper, we propose a novel method called RXFOOD for the fusion of features across different scales within the same modality branch and from different modality branches simultaneously in a unified attention mechanism. An Energy Exchange Module is designed for the interaction of each feature map's energy matrix, who reflects the inter-relationship of different positions and different channels inside a feature map. The RXFOOD method can be easily incorporated to any dual-branch encoder-decoder network as a plug-in module, and help the original backbone network better focus on important positions and channels for object of interest detection. Experimental results on RGB-NIR salient object detection, RGB-D salient object detection, and RGB-Frequency image manipulation detection demonstrate the clear effectiveness of the proposed RXFOOD. 
\end{abstract}


%
\IEEEpeerreviewmaketitle

\section{Introduction}
\IEEEPARstart{W}{ith} the rapid development of different types of sensors, \eg RGB camera, Depth, Near-infrared (NIR) \etc, human beings are able to acquire various data from the surrounding world now. Each data source provides us a unique perspective of real world, for example, RGB camera delivers colored image by capturing light in red, green, and blue wavelengths, Depth sensor determines the range of an object or a surface to the sensor~\cite{zhou2021ccafnet}, NIR spectrum contains the material properties of samples that can be used for characterizing objects~\cite{song2020deep}. These different types of data are complementary to each other, and give us a more comprehensive understanding of captured scene for multimedia research and applications. 

The Fig.~\ref{fig:motivation}(a) shows three multimedia applications for the RGB-X Object of Interest (OoI) Detection task, where object of interest can be anything that we want to distinguish from others. (1) The RGB image is sometimes vulnerable to light condition, thus does not contain enough useful information in dark environment, while near-infrared (NIR) can still satisfy visual recognition purpose for salient object detection. (2) In addition, the depth image could assist the RGB images to localize the salient object by giving the extra depth cue. (3) It might be hard to distinguish the tampered region relying only on RGB information, but the Frequency information could be beneficial for the image manipulation detection task. Based on these facts, recent deep learning based neural networks take advantage of multiple data modalities in order to overcome the bias of single data source~\cite{song2020deep, liu2020S2MA, wang2022objectformer}. In this paper, we denote these deep neural networks for RGB-X OoI as RGB-X encoder-decoder networks, where X represents any type of modality data that is complementary to RGB image and is able to serve as another input data.

The key point in RGB-X networks is how to fuse these two types of modality data for better performance. The fusion process can generally take place at three stages: early fusion~\cite{liu2019improved, song2017depth, peng2014rgbd}, intermediate fusion~\cite{chen2018progressively, liu2019improved, zeng2019deep}, and late fusion~\cite{cheng2017locality, han2017cnns}. Early fusion combines the RGB and X input together as one single input for network, the intermediate fusion aims to fuse feature representations in the middle layers, and the late fusion makes a unified prediction based on two raw predictions. Despite the simplicity of early fusion and late fusion, their performance is inferior due to the lack of information interactions~\cite{chen2018progressively}, so the intermediate fusion is a more popular and effective approach. Previous RGB-X related works typically apply naive fusion~\cite{song2020deep} (\eg, average or max fusion) or only focus on the feature fusion at the same scale(s)~\cite{liu2020S2MA}. Recent works take multi-scale features into consideration. Multi-scale features (or feature pyramid) have been demonstrated to be helpful for image perception~\cite{lin2017feature} by exploiting features from different scales. Current UNet~\cite{ronneberger2015u}-like architectures typically use multi-scale features as the guidance for corresponding scales in decoder. \textit{Unlike other works that consider each scale of the multi-scale features separately, this paper aims to design a fusion method that can fuse features across different scales within the same modality branch and from different modality branches simultaneously in a unified attention mechanism.} This module is a flexible plug-in method that can be applied into different RGB-X networks.


The attention mechanism can exploit the context relationship among different positions inside an  input~\cite{vaswani2017attention}, then produce a reformulated but more powerful representation for that input. Inspired by the attention mechanism, we propose the RXFOOD fusion method for the OoI detection, which is a flexible and compatible plug-in module with existing two-branch RGB-X encoder-decoder networks. The inter-relationship of different patches and channels inside an image is an important clue for finding the object of interest, we call this inter-relationship as \textbf{energy}, and introduce an Energy Exchange Module that enables the interaction of spatial and channel inter-relationship generated across different scales within the same modality branch and from different modality branches simultaneously. Experimental results show that the fusion process of RXFOOD can enhance features and improve the performance of original backbone network, and it performs better than other comparison fusion methods, as shown in Fig.~\ref{fig:motivation}(b). The contributions of our proposed RXFOOD method are summarized as follows.

\begin{itemize}
    \item We propose a novel method called RXFOOD as RGB-X fusion for the Object of Interest detection, which can fuse the RGB-X information by spatial and channel energy exchange across different scales within the same modality branch and from different modality branches simultaneously.   
    
    \item The proposed RXFOOD is compatible with existing two-branch RGB-X  encoder-decoder networks, which can be easily incorporated with backbone networks as a plug-in module.

    \item Our intensive experiments on three multimedia  tasks (RGB-NIR salient object detection, RGB-D salient object detection, and RGB-Frequency image manipulation detection) demonstrate that the performance of backbone networks could be improved when they are equipped with the proposed RXFOOD. 
    
\end{itemize}

\section{Related Work}
\subsection{RGB-X Object of Interest Detection}
In order to overcome the limitation of single data source, many recent deep learning based methods utilized the additional modality information to help the detection tasks, like depth images~\cite{chen2019multi,zhu2019pdnet,zhang2021bilateral,pang2020multi}, thermal infrared images~\cite{zhang2019rgb,cong2022does,tu2021multi,zhou2023lsnet,wu2023menet}, near-infrared (NIR) images~\cite{cong2018hscs,aslahishahri2021rgb,jin2022darkvisionnet}, \etc.  Different kinds of image modalities can provide different contributions for computer vision and  multimedia tasks. For example, the depth image could assist the RGB images by giving the extra depth information to improve the performance on salient object detection~\cite{liu2021learning,chen2020progressively,pang2020multi,zhou2021ccafnet} and scene parsing~\cite{zhou2022pgdenet}. NIR images contain rich shape and edge information that is useful for detection and segmentation~\cite{song2020deep,wang2013multi}, especially when the RGB image is not clear in low light environment. In the field of image manipulation detection, it is hard to distinguish authentic and tampered regions by using only RGB information. While researchers found that they may show different traces in the frequency domain~\cite{verdoliva2020media,wang2022objectformer}, so the frequency information is helpful in image manipulation detection task.  In this paper, we focus on the information fusion for RGB-X object of interest detection, where X can be any type of modality data.

\subsection{RGB-X Fusion} 
The fusion strategies for combining information from RGB branch and X branch can be roughly divided into three categories: early fusion~\cite{qu2017rgbd,fu2020jl,liu2019salient}, intermediate fusion~\cite{chen2019multi,zhao2019contrast,zhang2021bilateral,li2020icnet}, and late fusion~\cite{fang2014saliency,han2017cnns,zhu2019pdnet}. Despite the simplicity of early fusion and late fusion, recent state-of-the-art methods typically choose the intermediate representations for fusion due to its high performance on corresponding tasks~\cite{zeng2019deep,fan2020bbs,zhao2020suppress}. Multi-scale features are also important during the fusion process since they can exploit information from different scales. For example, \cite{hong2022reflection} fuses multi-scale features from RGB and NIR for reflection removal. \cite{li2020cross} proposes a cross-modal weighting strategy for fusion by using three interaction modules for low-, middle- and high-level features. \cite{fu2021siamese} proposes a siamese network for RGB-D saliency detection by using joint learning and densely cooperative fusion. However, previous fusion methods focus on either the calibrated fusion of each single-scale feature independently in a cascade manner across modality or the multi-level information fusion within single modality. While in this paper, we aim to propose a plug-in simultaneous fusion method for features across different scales within the same modality branch and from different modality branches in a unified framework. 


\begin{figure*}[ht]
	\begin{minipage}[b]{1\textwidth}
		\centering
		\includegraphics[width=0.85\textwidth]{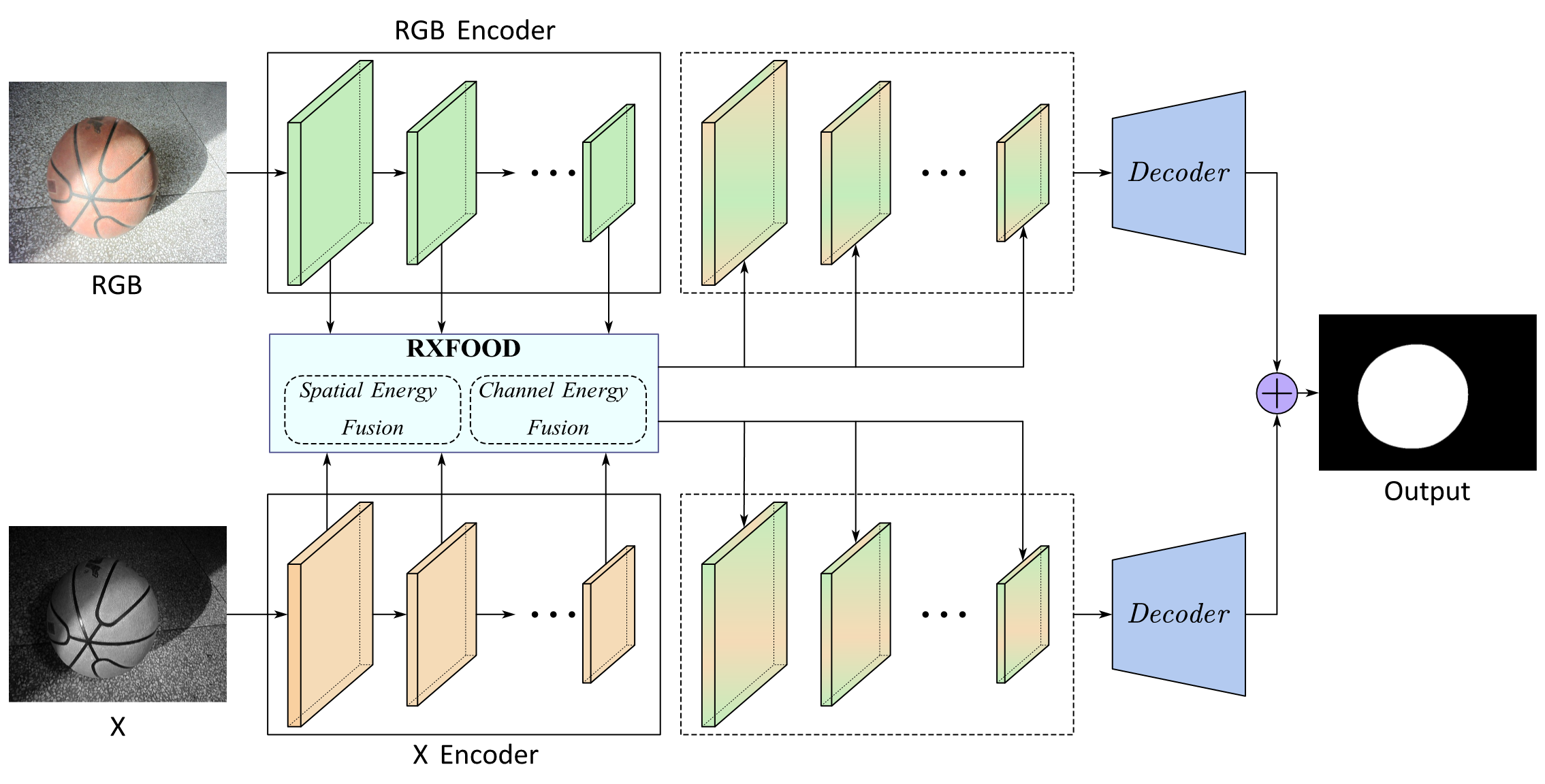}
	\end{minipage}	
	\caption{The pipeline of proposed RXFOOD method. RXFOOD takes multiscale features from different modality encoders as input, then outputs their fused counterparts. These fused multiscale features are fed into decoder for prediction. \textit{Note that the skip connections from multiscale features to decoder are not shown in this figure for simplicity.} } 
	\label{fig:pipeline} 
\end{figure*}

\subsection{Attention Mechanism}
Attention mechanism is introduced to deep learning inspired by human cognitive nature that people tend to pay attention to specific parts rather than entire scope when perceiving information~\cite{corbetta2002control}. By combing the attention module with a deep learning network, the model can focus on valuable information from input, just like human being, so it has been widely applied in various computer vision and multimedia   tasks~\cite{dang2020detection,gong2021mitigating,islam2020doa,isobe2020video,liu2019griddehazenet,wang2018non,fu2019dual}. For example, self-attention mechanism~\cite{vaswani2017attention} can draw global dependencies between a group of inputs, which can be applied in image generation~\cite{zhang2019self,parmar2018image}, image classification~\cite{wang2018non,li2019selective}. Dual attention~\cite{fu2019dual} exploit the long range correlation in spatial and channel dimension. Selective self-mutual attention(S$^2$MA)~\cite{liu2020S2MA} is designed to fuse the features from two different modalities. Our RXFOOD method is also inspired by attention mechanism, whose unique purpose is to fuse features across different scales within the same modality branch and from different modality branches via spatial and channel energy exchange.


\section{METHODOLOGY}\label{sec:MSCDA}
\subsection{RXFOOD Overview}
In this section, we introduce the proposed RXFOOD method. Our motivation is that the feature representations from different branches and different scales are complementary to each other, thus the information fusion can help reinforce each feature representation, and improve the network performance. By considering each feature map $F\in\mathbb{R}^{W\times H\times C}$ as a system contains $WH$ elements $\{\alpha_i\in\mathbb{R}^C, i=1,2,...,WH\}$, we can compute a spatial-wise similarity matrix with shape $WH\times WH$ that reflects the inter-relationship (similar or not) of each pair of elements inside the  system. In Markov Random Fields (MRF)~\cite{boykov2001fast,boykov2001interactive} and the Densely-connected MRF~\cite{zhang2016instance} research, the similarity between pairwise elements in MRF can be modeled as energy terms to describe system status. Inspired by their idea, we believe that the spatial-wise similarity matrix ($WH\times WH$) could reflect the system spatial status, so we name it as the \textbf{spatial energy} of a feature map based system in this paper. Similarly, we can compute the channel-wise similarity matrix ($C\times C$) as \textbf{channel energy} of a feature map based system, which represents the system channel status. These two energy matrices provide clues for finding important spatial positions and channels in object of interest detection, so we develop the proposed RXFOOD method based on energy fusion, which is different from the energy minimization in MRF related works. 

As illustrated in Fig.~\ref{fig:pipeline}, an encoder-decoder backbone network is applied to RGB input and X input respectively with separate parameters. The encoder stage progressively downsamples the resolution of input data, and map input into feature representations of different scales. The decoder stage takes the feature representations from encoder as input, then make final prediction. Skip connections between encoder and decoder are commonly used approach to make use of multiscale features. Our plug-in RXFOOD method takes multiscale features from two encoders as input, then outputs their fused counterpart features. For example, given the multiscale features extracted by a backbone network from RGB image as $\{F_{rgb}^1, F_{rgb}^2, ..., F_{rgb}^n\}$, and from X image as $\{F_{x}^1, F_{x}^2, ..., F_{x}^n\}$, where $n$ is the number of scales, and each feature has its own perceptive field. Then the input for our RXFOOD method are these features from different branches and different scales, and the output after fusion are their counterparts $\{\hat{F}_{rgb}^1, \hat{F}_{rgb}^2, ..., \hat{F}_{rgb}^n\}$ and $\{\hat{F}_{x}^1, \hat{F}_{x}^2, ..., \hat{F}_{x}^n\}$. These fused features are fed into decoder for final prediction.

There are two components in our RXFOOD method: RGB-X spatial energy fusion, RGB-X channel energy fusion. Next, we will explain these two components in details.


\subsection{RGB-X Spatial Energy Fusion}
\textbf{Spatial Energy Computation:}
The spatial energy matrix is a by-product of spatial attention mechanism. Given an input feature map $F\in\mathbb{R}^{W\times H\times C}$ from the encoder, spatial attention exploits the contextual clues among different spatial positions, thus helps the network focus on important regions in object of interest detection. Fig.\ref{fig:spatial_energy} shows the general pipeline of spatial attention, first we construct the query $Q\in\mathbb{R}^{W\times H\times d}$, key $K\in\mathbb{R}^{W\times H\times d}$ and value $V\in\mathbb{R}^{W\times H\times C}$ from input feature map $F$ by Conv layers, where $d$ is the number of hidden channels for query and key. Then, we reshape them into query $Q\in\mathbb{R}^{N \times d}$, key $K\in\mathbb{R}^{N \times d}$ and value $V\in\mathbb{R}^{N \times C}$, where $N=W\times H$. We then compute the \textbf{spatial energy} matrix by the following scaled  dot-product~\cite{vaswani2017attention}: 

\begin{equation}\label{equ:energy_spatial}
    \varepsilon=\frac{Q \otimes K^\top}{\sqrt{d}},  
\end{equation} 
where $^\top$ is the transpose operation. This energy matrix reflects the long range  inter-dependency clues among all spatial positions. Traditionally, the final output of spatial attention module is computed by the following equation:
\begin{equation}\label{equ:attention_spatial}
AS=\alpha\cdot \mathrm{Reshape}(\mathrm{Softmax}(\varepsilon)\otimes V) \oplus F, 
\end{equation}
where $\alpha$ is a parameter controls the weight of attention, which is a learnable parameter with initial value of 0 in our method. $\mathrm{Reshape}$ means to recover the feature to original shape. $\otimes$ denotes matrix multiplication, and $\oplus$ denotes element-wise summation. We denote $AS$ as spatial attention aware feature. The procedure for computing $\varepsilon$ is marked as hidden step $h_1$, and the remaining computation for $AS$ is marked as hidden step $h_2$, as shown in Fig.\ref{fig:spatial_energy}.

\textbf{Spatial Energy Fusion: }
As shown in Fig.~\ref{fig:spatial_eem}, suppose we have the multiscale features $\{F_{rgb}^1, F_{rgb}^2, ..., F_{rgb}^n\}$ from RGB branch encoder and $\{F_x^1, F_x^2, ..., F_x^n\}$ from X branch encoder, where $n$ is the total number of scales. Features with lower superscript come from previous layers, so they have larger spatial resolution than those with higher superscript. Our spatial energy fusion method takes these features as input, then produces spatial attention aware fused features $\{AS_{rgb}^1, AS_{rgb}^2, ..., AS_{rgb}^n\}$ and $\{AS_x^1, AS_x^2, ..., AS_x^n\}$. For each feature $F_{z}^i$, where superscript $i\in\{1,2,...,n\}$ represents scale and subscript $z\in\{rgb, x\}$ represents branch, we can compute its corresponding spatial energy matrix $\varepsilon_{z}^i$ by changing Eq.~\ref{equ:energy_spatial} to:
\begin{equation}\label{equ:energy_spatial_multi}
    \varepsilon_{z}^i=\frac{Q_{z}^{i} \otimes (K_{z}^i)^\top}{\sqrt{d}},  
\end{equation} 
where $Q_{z}^i$ and $K_{z}^i$ are query and key computed from $F_z^i$ following the hidden step $h_1$ as we described above in spatial energy computation. We design an \textit{Spatial Energy Exchange Module} for the fusion of spatial energy matrices from all input, which takes $\{\varepsilon_{rgb}^1, \varepsilon_{rgb}^2, ..., \varepsilon_{rgb}^n\}$ and $\{\varepsilon_{x}^1, \varepsilon_{x}^2, ..., \varepsilon_{x}^n\}$ as input then output fused energy matrices $\{\hat{\varepsilon}_{rgb}^1, \hat{\varepsilon}_{rgb}^2, ... \hat{\varepsilon}_{rgb}^n\}$ and $\{\hat{\varepsilon}_{x}^1, \hat{\varepsilon}_{x}^2, ... \hat{\varepsilon}_{x}^n\}$. These fused spatial energy matrices are then utilized to compute the spatial attention aware fused features, by modifying Eq.~\ref{equ:attention_spatial} as: 

\begin{figure*}[t]
    \centering
    \subfigure[Spatial energy computation]{
         \centering
         \includegraphics[width=0.2\textwidth]{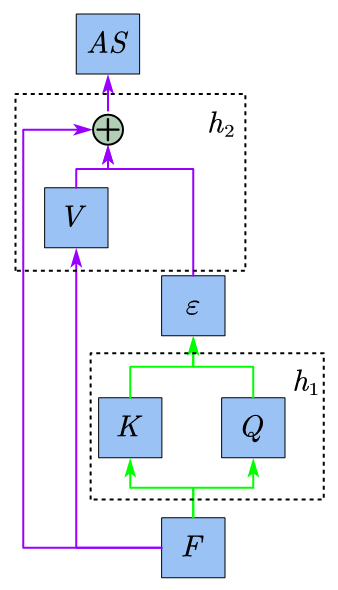}
         \label{fig:spatial_energy}
     }
     \hspace{3em}
     \subfigure[Spatial energy fusion]{
         \centering
         \includegraphics[width=0.6\textwidth]{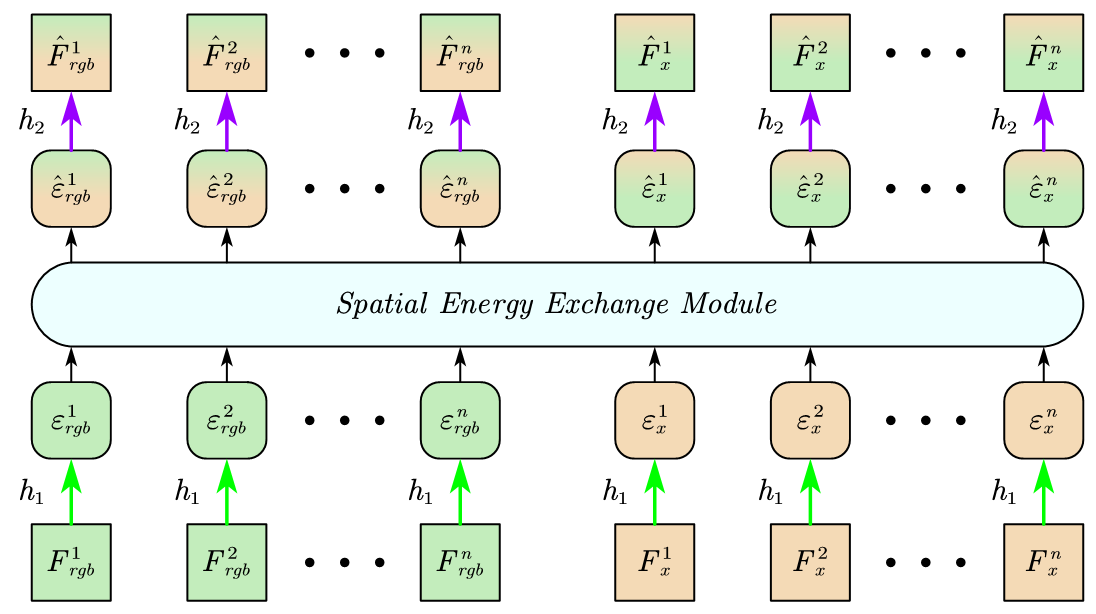}
         \label{fig:spatial_eem}
     }
    \caption{Illustration of  
    (a) Spatial energy computation, including $h_1$ and $h_2$ steps, (b) Our spatial energy fusion method.} 
    \label{fig:spatial_energy_fusion} 
\end{figure*}

\begin{equation}\label{equ:attention_spatial_multi}
\begin{split}
AS^i_{z}=\alpha\cdot \mathrm{Reshape}(\mathrm{Softmax}(\hat{\varepsilon}^i_{z})\otimes V^i_{z}) \oplus F^i_{z}, \\
i\in\{1,...,n\}, z\in\{rgb, x\}.
\end{split}
\end{equation}

Specifically, Fig.~\ref{fig:cross} shows the detailed structure of our \textit{Spatial Energy Exchange Module}. As the input spatial energy matrices have different size, first we need to re-scale these energy matrices to the same size. We set the maximum size of all energy matrices as a target size $S$, which is the size of $\varepsilon_{rgb}^1$ and $\varepsilon_x^{1}$ in our case. Then upsample all the other energy matrices to the target size $S$ except $\varepsilon_{rgb}^1$ and $\varepsilon_x^{1}$. We concatenate the upsampled energy matrices together and feed them into a 1$\times$1 Conv layer for information interaction. After information interaction, according to the indexes of those upsampled energy matrices, we downsample corresponding fused energy matrices to the original size. The whole process of our \textit{Spatial Energy Exchange Module} can be represented as:
\begin{equation}\label{equ:energy_fusion_spatial}
\begin{split}
    \{\hat{\varepsilon}_{rgb}^1,...,\hat{\varepsilon}_{rgb}^n\}, \{\hat{\varepsilon}_{x}^1,...,\hat{\varepsilon}_{x}^n\} = \qquad \qquad \qquad\\
    SEEM(\{\varepsilon_{rgb}^1,...,\varepsilon_{rgb}^n\}, \{\varepsilon_{x}^1,...,\varepsilon_{x}^n\}),
\end{split}
\end{equation}
where $n$ is the number of scales, \textit{SEEM} represents \textit{Spatial Energy Exchange Module}.

\subsection{RGB-X Channel Energy Fusion}
\textbf{Channel Energy Computation: }
While spatial attention explores the inter-relationship among spatial positions of input, channel attention can help network emphasize channels that are important for object of interest detection, by exploiting inter-dependencies among different channels.

Similarly, the channel energy matrix is the by-product of channel attention, the input feature $F\in\mathbb{R}^{W\times H\times C}$ is firstly converted to $F'\in\mathbb{R}^{W\times H\times c}$ by a 1$\times$1 Conv layer, where $c$ is a pre-defined number smaller than $C$ in order to reduce computational complexity. Then $F'$ is reshaped to $f\in\mathbb{R}^{N \times c}$, this $f$ is directly used as $Q, K, V$ to compute the \textbf{channel energy} matrix by: 
\begin{equation}\label{equ:energy_channel} 
\xi = \frac{f^\top \otimes f}{\sqrt{N}}.  
\end{equation}

Then the output of this channel attention module is computed  by: 
\begin{equation}\label{equ:attention_channel} 
AC=\beta\cdot \mathrm{Conv}(\mathrm{Reshape}(\mathrm{Softmax}(\xi)\otimes f)) \oplus F, 
\end{equation}
where $\beta$ is also a learnable parameter with initial value of 0. $\mathrm{Conv}$ is another 1$\times$1 Conv layer for recovering number of channels from $c$ to $C$. We denote $AC$ as channel attention aware feature. 

\begin{figure}[t]
	\centering
	\includegraphics[width=1\linewidth]{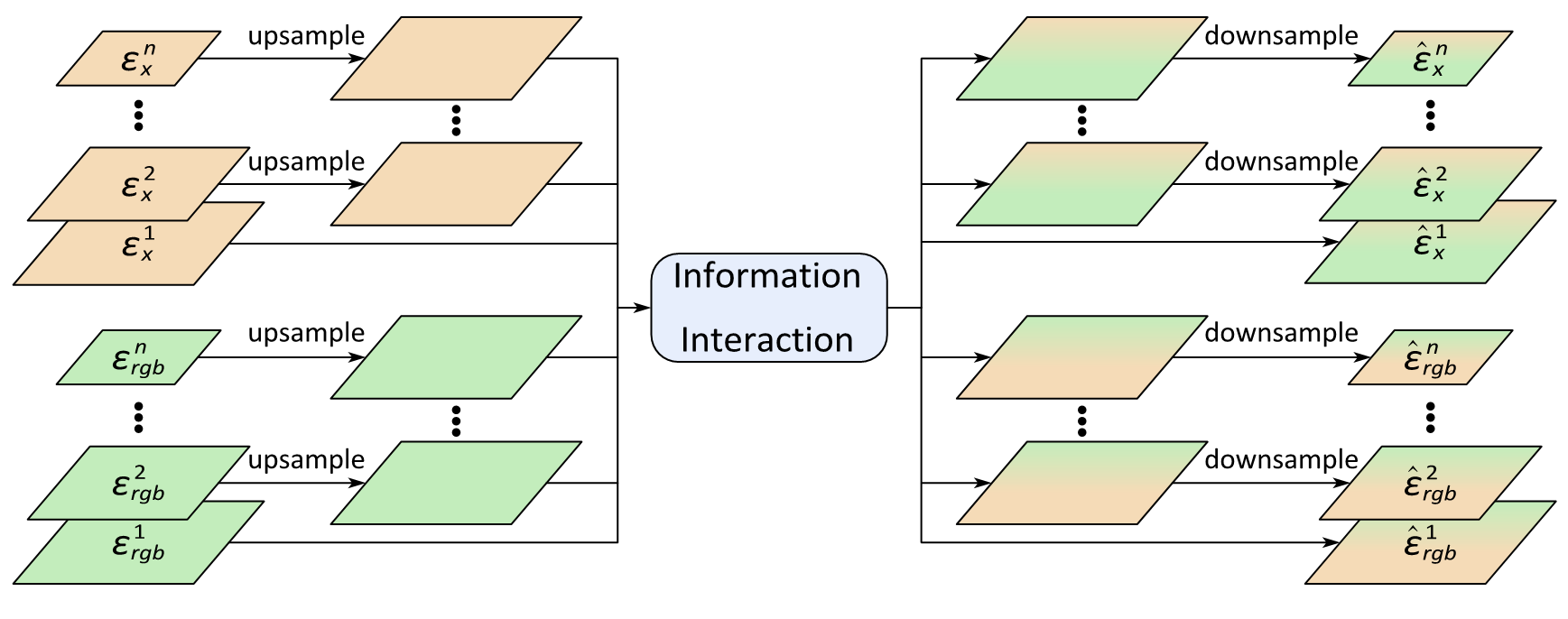}
	\caption{Illustration of the proposed \textit{Spatial Energy Exchange Module} as shown in Eq.~\ref{equ:energy_fusion_spatial}. All input spatial energy matrices obtain their fused counterparts after this module.} 
	\label{fig:cross} 
\end{figure}

\textbf{Channel Energy Fusion:}
Just the same way as in spatial energy fusion, given $F_{z}^i$ from input of RXFOOD, we first compute its corresponding channel energy matrix by changing Eq.~\ref{equ:energy_channel} to:
\begin{equation}\label{equ:energy_channel_multi} 
\xi_z^i = \frac{(f_z^i)^\top \otimes f_z^i}{\sqrt{N}},  
\end{equation}
where $f_z^i\in \mathbb{R}^{N\times c}$ is generated from $F_{z}^i$. Then we can feed these channel energy matrices into the \textit{Channel Energy Exchange Module} to get fused output: 
\begin{equation}\label{equ:energy_fusion_channel}
    \begin{split}
    \{\hat{\xi}_{rgb}^1,...,\hat{\xi}_{rgb}^n\}, \{\hat{\xi}_{x}^1,...,\hat{\xi}_{x}^n\} = \qquad \qquad \qquad\\
    CEEM(\{\xi_{rgb}^1,...,\xi_{rgb}^n\}, \{\xi_{x}^1,...,\xi_{x}^n\}),
    \end{split}
\end{equation}
where \textit{CEEM} has the same structure as \textit{SEEM} as shown in Fig.~\ref{fig:cross} except the input. These fused channel energy matrices are then applied to compute the channel attention aware features by modifying Eq.~\ref{equ:attention_channel} to:
\begin{equation}\label{equ:attention_channel_multi}
\begin{split}
AC_{z}^i=\beta\cdot \mathrm{Conv}(\mathrm{Reshape}(\mathrm{Softmax}(\hat{\xi}_{z}^i)\otimes f_{z}^i)) \oplus F_{z}^i, \\
i\in\{1,...,n\}, z\in\{rgb, x\}.
\end{split}
\end{equation}

After obtaining the spatial attention aware feature $AS_{z}^i$ and channel attention aware feature $AC_{z}^i$ for each input feature $F_z^i$, its fused counterpart feature $\hat{F}_{z}^i$ is computed as:
\begin{equation}\label{equ:spatial_plus_channel} 
\hat{F}_{z}^i = AS_{z}^i \oplus AC_{z}^i, i\in\{1,...,n\}, z\in\{rgb, x\}.
\end{equation}

These fused counterpart features $\hat{F}_{z}^i$ will be used in decoder network to replace the original features $F_{z}^i$ for final prediction. With the help of the \textit{Spatial Energy Exchange Module} and \textit{Channel Energy Exchange Module}, our RXFOOD method can fuse features across different scales within the same modality branch and from different modality branches simultaneously, and help the backbone network focus more on specific positions and channels that are related to object of interest detection. 

\subsection{Loss Function}
We consider the object of interest detection problem as a pixel-level binary classification problem (either OoI or not), and employ the Binary Cross Entropy (BCE) loss to ensure the correctness of prediction. Let us define the predicted mask as $X\in\mathbb{R}^{W\times H}$ and the  target/ground truth as $Y\in\mathbb{R}^{W\times H}$, then the BCE loss is defined as:
\begin{equation}\label{equ:bce}
    \mathcal{L}_{BCE} = -\sum_{w=1}^W\sum_{h=1}^H y_{wh}\log{x_{wh}} + (1-y_{wh})\log{(1-x_{wh})}. 
\end{equation}

\section{Experiments} 
This section evaluates several methods  on three RGB-X Object of Interest detection tasks, \ie, RGB-N(IR) salient object detection, RGB-D salient object detection and RGB-F(requency) image manipulation detection. 

\subsection{Experimental Setting}
\textbf{X branch input:} RGB-N aims to detect salient object by synchronized RGB images and near-infrared images, so the input for X branch is NIR image. While RGB-D aims to detect salient object by synchronized RGB images and depth images, the input for X branch is depth image. For RGB-F image manipulation detection task, we extract the frequency information from RGB images as the input for X branch, the frequency extractor is Frequency-aware Decomposition (FAD) proposed in ~\cite{qian2020thinking}.

\textbf{Datasets:} For the RGB-N task, we use the public  RGBN-SOD dataset~\cite{song2020deep} to test our RXFOOD method. This dataset contains 780 synchronized RGB and NIR image pairs, in which 390 pairs are used for training, 78 pairs for validation, and 312 pairs for testing. For the RGB-D task, we use the public ReDWeb-S dataset~\cite{liu2021learning} with 3,179 RGB and depth image pairs, which is divided into a training set with 2,179 image pairs and a testing set with the remaining 1,000 image pairs. For the  RGB-F task, we use the public CASIA dataset~\cite{Dong2013casia} to test RXFOOD, the training set is CASIA2.0 with 5,123 tampered images and the testing set is CASIA1.0 with 920 tampered images.

\textbf{Evaluation metrics:} For RGB-N and RGB-D salient object detection tasks, we use the same metrics Smeasure~\cite{Smeasure}, Emeasure~\cite{Emeasure}, Fmeasure~\cite{Fmeasure}, maxF score and MAE~\cite{MAE} as in~\cite{song2020deep,liu2020S2MA}. For RGB-F image manipulation detection task, we use the area under the ROC curve (AUC) and the maxF score to evaluate the performance, same as the evaluation in~\cite{zhou2018learning, salloum2018image}. For the MAE metric, the lower the better, while for the others, the higher the better.

\textbf{Backbone networks:} As a plug-in, our RXFOOD method can be easily incorporated with other encoder-decoder based networks. For the  RGB-N task, we choose its state-of-the-art Two-Branch Network (TBNet)~\cite{song2020deep} as our backbone. Specifically, BASNet~\cite{qin2019basnet} and CPDNet~\cite{wu2019cascaded} are employed to implement each branch of TBNet following its original  setting~\cite{song2020deep}, denoted as TBNet$_\mathrm{BAS}$ and TBNet$_\mathrm{CPD}$ respectively. While for the RGB-D task, we first adopt the existing two-branch network S$^2$MANet~\cite{liu2020S2MA} as a backbone network, then TBNet$_\mathrm{BAS}$ and the two-branch Feature Pyramid Networks (FPN)~\cite{kirillov2019panoptic} are employed as other backbone networks. Note that S$^2$MANet refers to the whole network in~\cite{liu2020S2MA}, while S$^2$MA indicates only the Selective Self-Mutual Attention module in their work.  As for RGB-F task, we employ TBNet$_\mathrm{BAS}$ and the two-branch FPN~\cite{kirillov2019panoptic} as backbone networks. For TBNet$_\mathrm{BAS}$,  TBNet$_\mathrm{CPD}$, and S$^2$MANet, we use features from the last three different scales (n=3) of encoder as input for fusion. For FPN, we use all its four scales  (n=4) of encoder as input for fusion. This setting is to show  that the proposed RXFOOD method is a flexible plug-in to be easily incorporated into different encoder-decoder based backbone networks with the customized number of scales that the user wants.

\textbf{Comparison methods:} We choose three intermediate fusion methods for comparison, including simple average fusion (AVG), max fusion (MAX), and selective self-mutual attention (S$^2$MA) based fusion~\cite{liu2020S2MA}. AVG fusion simply computes the average of input features for fusion, while MAX fusion computes the maximum activation of input features. We apply these two simple fusion methods on same scale features and on multiscales independently. S$^2$MA is a specifically designed attention for fusion purpose, and we apply it to the last-layer features of encoder by following the same setting in their original paper~\cite{liu2020S2MA}. \textit{For each task, we only compare these fusion methods on the strongest backbone network (for that task) with the highest maxF score}.


\subsection{RGB-N Salient Object Detection}
\begin{figure*}
     \centering     
     \subfigure[Qualitative examples on RGB-NIR salient object detection. Top: similar background; Bottom: dark  condition.]{
         \centering
         \includegraphics[width=0.95\textwidth]{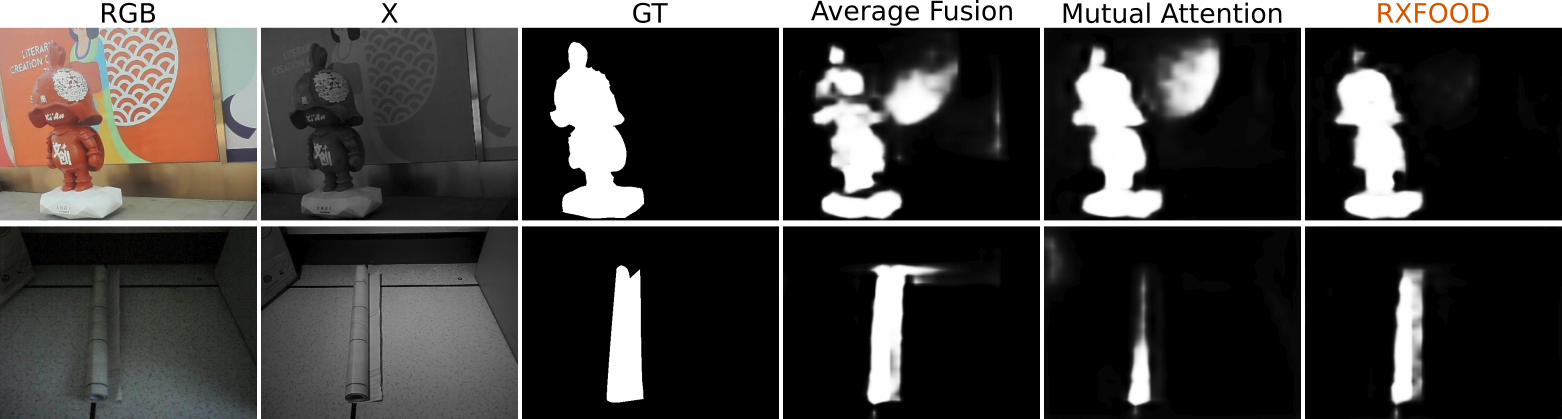}
         \label{fig:mssod_result}
     }
     \subfigure[Qualitative examples on RGB-D salient object detection. Top: complex background; Bottom: similar background. ]{
         \centering
         \includegraphics[width=0.95\textwidth]{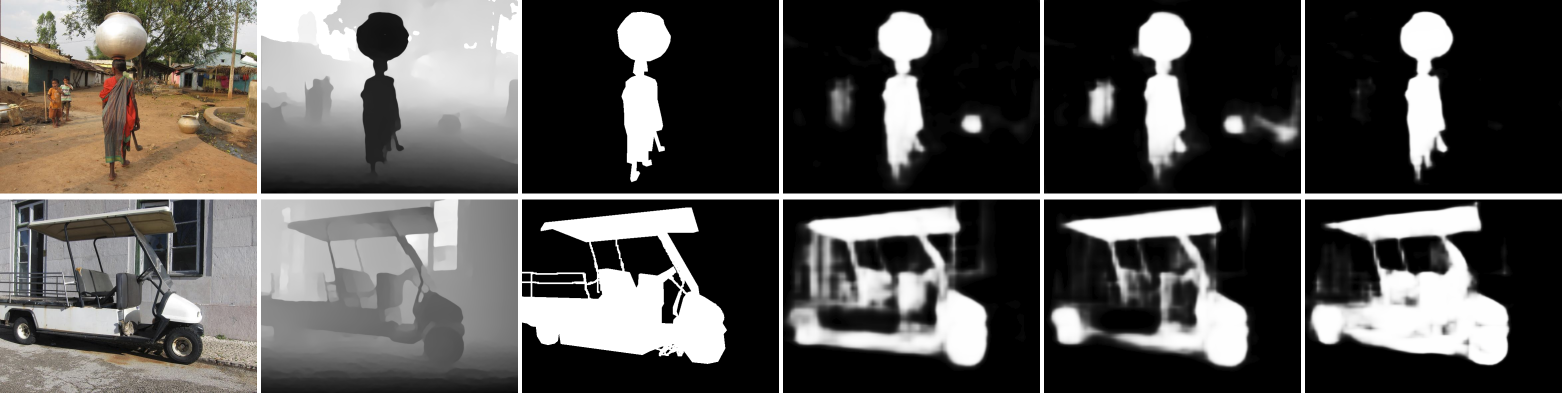}
         \label{fig:rdsod_result}
     }
     \subfigure[Qualitative examples on RGB-Frequency image manipulation detection. Top: splicing; Bottom: copy-move.]{
         \centering
         \includegraphics[width=0.95\textwidth]{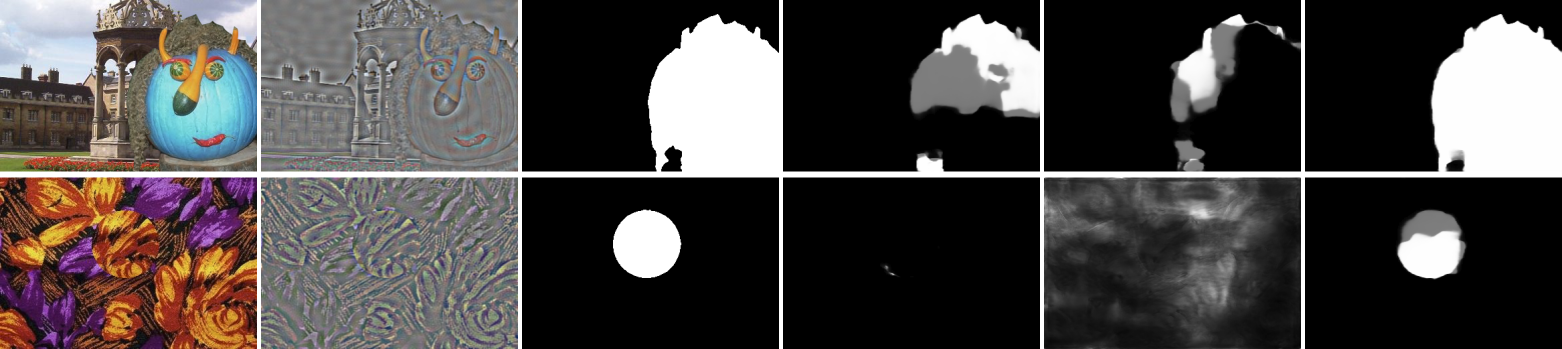}
         \label{fig:imd_results}
     }
    \caption{Qualitative results on three RGB-X Object of Interest detection  tasks. From left to right: RGB images, X input images, ground truth, average fusion results, S$^2$MA~\cite{liu2020S2MA} results, and our RXFOOD results.}
    \label{fig:qualitative}
\end{figure*}

We show quantitative experimental results of RGB-N task in Table.~\ref{tab:mssod}. As we can see from the table, by adding the proposed RXFOOD fusion method to TBNet$_\mathrm{BAS}$, the performance in all metrics are improved. Specifically, there is a large improvement of 5.9\% in Fmeasure, and 5\% in maxF. TBNet$_\mathrm{CPD}$ is a more powerful backbone network, its performance is higher than TBNet$_\mathrm{BAS}$ on all metrics. When applying AVG to TBNet$_\mathrm{CPD}$, only Fmeasure and MAE performance are slightly improved by 0.88\% and 0.1\% respectively, the performance in other metrics are even damaged, this phenomenon also happens in MAX. This is because features from different modalities have information disparity, they cannot be directly combined together. These simple fusion methods cannot address this issue. S$^2$MA is beneficial for original backbone network, it improves the original backbone network performance on all metrics by over 0.5\%.  But its performance is still inferior to our RXFOOD method, which obtains the best scores on all metrics. We display two RGB-N qualitative examples in Fig.~\ref{fig:mssod_result} under similar background condition (top) and dark condition (bottom) respectively. We can see our RXFOOD performs better than other fusion methods under these challenging conditions. 

\begin{table}[ht]
    \caption{Experimental result comparison on the RGB-N salient object detection task.}
	\label{tab:mssod}
	\centering
	\resizebox{\columnwidth}{!}
	{
	\begin{tabular}{l|c|c|c|c|c}
		\toprule
		  Method & Smeasure & Emeasure & Fmeasure & maxF & MAE$\downarrow$  \\
		\midrule
	      TBNet$_\mathrm{BAS}$\cite{song2020deep} & 0.8783 & 0.9122 & 0.8231 & 0.8665 & 0.0496 \\
            TBNet$_\mathrm{BAS}$+RXFOOD & 0.9180 & 0.9448 & 0.8821 & 0.9165 & 0.0321 \\
            \midrule
            TBNet$_\mathrm{CPD}$~\cite{song2020deep} & 0.9152 & 0.9486 & 0.8741 & 0.9107 & 0.0337 \\
            TBNet$_\mathrm{CPD}$+AVG & 0.9120 & 0.9482 & 0.8829 & 0.9064 & 0.0327 \\
            TBNet$_\mathrm{CPD}$+MAX & 0.9051 & 0.9380 & 0.8666 & 0.8961 & 0.0365 \\
            TBNet$_\mathrm{CPD}$+S$^2$MA~\cite{liu2020S2MA} & 0.9214 & 0.9555 & 0.8888 & 0.9171 & 0.0284 \\
            TBNet$_\mathrm{CPD}$+RXFOOD & \textbf{0.9232} & \textbf{0.9578} & \textbf{0.8953} & \textbf{0.9212} & \textbf{0.0267} \\
		\bottomrule
	\end{tabular}
	}
\end{table}

\subsection{RGB-D Salient Object Detection}
The quantitative experimental results of RGB-D task are shown in Table.~\ref{tab:rdsod}. Our RXFOOD method can still improve the performance of backbone networks in this task. For TBNet$_\mathrm{BAS}$, the Emeasure is improved from 0.7447 to 0.7572, Fmeasure is improved from 0.6459 to 0.6601. For S$^2$MANet~\cite{liu2020S2MA}, we can further improve its performance by adding our RXFOOD fusion method to this dual-branch network. Specifically, the maxF score increases from 0.7302 to 0.7364, and the MAE drops from 0.1346 to 0.1298. The improvement by RXFOOD is more significant in FPN backbone network, with 3.99\% in Emeasure, 4.84\% in Fmeasure, and 1.57\% drop in MAE. Moreover, we can still observe that AVG and MAX fusions are prejudicial to the FPN backbone network in metrics Smeasure and maxF, which demonstrate that these two simple fusion methods are still not good enough in RGB-D applications. S$^2$MA does not perform well when applying to FPN, even worse than AVG and MAX fusion. This may due to that mutual attention is designed to fuse information only in the bottleneck stage, but the information in FPN is distributed in feature pyramid. On the contrary, our RXFOOD is still compatible with FPN as a plug-in module. Fig.~\ref{fig:rdsod_result} shows two examples on RGB-D task, comparing with other fusion methods, our RXFOOD can make more precise predictions.  

\begin{table}[ht]
    \caption{Experimental result comparison on the RGB-D salient object detection task.}
	\label{tab:rdsod}
	\centering
	\resizebox{\columnwidth}{!}
	{
	\begin{tabular}{l|c|c|c|c|c}
		\toprule
		  Method & Smeasure & Emeasure & Fmeasure & maxF & MAE$\downarrow$  \\
            \midrule
		TBNet$_\mathrm{BAS}$\cite{song2020deep} & 0.7506 & 0.7447 & 0.6459 & 0.7411 & 0.1367 \\
            TBNet$_\mathrm{BAS}$+RXFOOD & 0.7629 & 0.7572 & 0.6601 & 0.7483 & 0.1284 \\
            \midrule
            S$^2$MANet~\cite{liu2020S2MA} & 0.7539 & 0.7704 & 0.6955 & 0.7302 & 0.1346 \\
            S$^2$MANet+RXFOOD & 0.7595 & 0.7781 & 0.7023 & 0.7364 & 0.1298 \\
            \midrule
            FPN~\cite{kirillov2019panoptic} & 0.7716 & 0.7640 & 0.6791 & 0.7563 & 0.1271 \\
            FPN+AVG & 0.7697 & 0.7712 & 0.6943 & 0.7531 & 0.1234 \\
            FPN+MAX & 0.7665 & 0.7861 & 0.7091 & 0.7520 & 0.1216 \\
            FPN+S$^2$MA~\cite{liu2020S2MA} & 0.7630 & 0.7671 & 0.6870 & 0.7432 & 0.1260 \\
            FPN+RXFOOD & \textbf{0.7735} & \textbf{0.8039} & \textbf{0.7275} & \textbf{0.7610} & \textbf{0.1114} \\
		\bottomrule
	\end{tabular}
	}
\end{table}

\subsection{RGB-F Image Manipulation Detection}
Table.~\ref{tab:manipulation} shows experimental results of RGB-F task. When applying RXFOOD to FPN, the maxF score is improved from 0.3796 to 0.4019, and AUC is improved from 0.7601 to 0.7884. The TBNet$_\mathrm{BAS}$ obtains higher maxF score than FPN, but lower AUC score, because these two metrics are sometimes not positively correlated, where the similar mismatch was reported in the previous work~\cite{wang2022objectformer}. The AVG can improve maxF from 0.3960 to 0.4114, AUC from 0.6974 to 0.7022, while MAX can only improve the AUC score to 0.7024. S$^2$MA can improve the maxF score to 0.4127 and AUC score to 0.7085. When applying our RXFOOD method to TBNet$_\mathrm{BAS}$, we get the best maxF 0.4380 and AUC 0.7273, comparing with other fusion methods. There are two types of manipulation in testing set, splicing and copy-move, we show one qualitative example for each manipulation type respectively in Fig.~\ref{fig:imd_results}. The top row is an splicing example where RXFOOD performs much better than other two fusion methods. The bottom row is a difficult copy-move case for all fusion methods, but our RXFOOD can still find the location of indistinguishable tampered region.

\begin{table}[ht]
    \caption{Experimental result comparison on the RGB-F image manipulation detection task.}
	\label{tab:manipulation}
	\centering
	\resizebox{0.7\columnwidth}{!}
	{
	\begin{tabular}{l|c|c}
		\toprule
		  Method & maxF & AUC  \\
		\midrule
		FPN~\cite{kirillov2019panoptic} & 0.3796 & 0.7601 \\
            FPN+RXFOOD & 0.4019 & \textbf{0.7884} \\
            \midrule
            TBNet$_\mathrm{BAS}$~\cite{song2020deep} & 0.3960 & 0.6974  \\
            TBNet$_\mathrm{BAS}$+AVG & 0.4114 & 0.7022 \\
            TBNet$_\mathrm{BAS}$+MAX & 0.3901 & 0.7024 \\
            TBNet$_\mathrm{BAS}$+S$^2$MA~\cite{liu2020S2MA} &  0.4127 & 0.7085 \\
            TBNet$_\mathrm{BAS}$+RXFOOD & \textbf{0.4380} & 0.7273\\
		\bottomrule
	\end{tabular}
	}
\end{table}

\subsection{Ablation Study}
In order to study the benefits of fusion for multiscales and different modality branches, we conduct ablation study on RGB-N salient object detection task by using TBNet$_\mathrm{CPD}$ as backbone network. Our RXFOOD can be modified to accept single scale features $\{F_{rgb}, F_x\}$ as input, then we apply it to the last layer features from encoder as in S$^2$MANet~\cite{liu2020S2MA}, and denote it as SF (single scale fusion). We also apply SF to multiscale features $\{F_{rgb}^1, F_x^1\}, \{F_{rgb}^2, F_x^2\}, ..., \{F_{rgb}^n, F_x^n\}$ independently, and denote it as MSF (multiple single scale fusion). Besides, S$^2$MA~\cite{liu2020S2MA} is also applied to multiscale features independently in order to study its performance on multiscale fusion, which is denoted as MMA (multiple mutual attention). Table.~\ref{tab:ablation} shows experimental results for ablation study. As we can see from this table, although S$^2$MA is able to improve the original TBNet$_\mathrm{CPD}$, its multiscale version MMA performs even worse than original TBNet$_\mathrm{CPD}$. 
This indicates Mutual Attention is not suitable for the fusion of multiscale features. The TBNet$_\mathrm{CPD}$+SF slightly improves the performance of original TBNet$_\mathrm{CPD}$, showing that our energy fusion idea is useful even in single scale condition. By extending SF to multiscale, the performance is further improved, and TBNet$_\mathrm{CPD}$+MSF is competitive with TBNet$_\mathrm{CPD}$+S$^2$MA, with higher performance on S-Mearure, maxF and MAE. Finally, our RXFOOD method achieves the best score on all metrics, showing that the fusion for features across different scales within the same modality branch and from different modality branches is an effective way for improving network performance.

\begin{table}[ht]
    \caption{Ablation study on RGB-N task by using TBNet$_\mathrm{CPD}$.}
	\label{tab:ablation}
	\centering
	\resizebox{\columnwidth}{!}
	{
	\begin{tabular}{l|c|c|c|c|c}
		\toprule
		  Method & Smeasure & Emeasure & Fmeasure & maxF & MAE$\downarrow$  \\
            \midrule
            TBNet$_\mathrm{CPD}$~\cite{song2020deep} & 0.9152 & 0.9486 & 0.8741 & 0.9107 & 0.0337 \\
            TBNet$_\mathrm{CPD}$+S$^2$MA~\cite{liu2020S2MA} & 0.9214 & 0.9555 & 0.8888 & 0.9171 & 0.0284 \\
            TBNet$_\mathrm{CPD}$+MMA & 0.9045 & 0.9398 & 0.8624 & 0.8940 & 0.0327 \\
            \midrule
            TBNet$_\mathrm{CPD}$+SF & 0.9169 & 0.9494 & 0.8778 & 0.9135 & 0.0322 \\
            TBNet$_\mathrm{CPD}$+MSF & 0.9220 & 0.9543 & 0.8851 & 0.9178 & 0.0277 \\
            TBNet$_\mathrm{CPD}$+RXFOOD & \textbf{0.9232} & \textbf{0.9578} & \textbf{0.8953} & \textbf{0.9212} & \textbf{0.0267} \\
		\bottomrule
	\end{tabular}
	}
\end{table}

\subsection{Inference Time}
Table~\ref{tab:time} shows the average inference time on each image of the whole testing set on the task of RGB-N salient object detection. All experiments are tested on a single NVIDIA 3090 GPU card. As we can see from the table, the inference time of TBNet$_\mathrm{BAS}$ and TBNet$_\mathrm{CPD}$ for a 256$\times$256$\times$3 color image are 0.035s and 0.023s respectively. After applying RXFOOD, the inference time is only increased by 0.002s on TBNet$_\mathrm{BAS}$ and 0.004s on TBNet$_\mathrm{CPD}$. While the time increase of AVG and MAX on TBNet$_\mathrm{CPD}$ is 0.001s, and the time increase of S$^2$MA on TBNet$_\mathrm{CPD}$ is 0.002s. 
Comparing with the inference time of backbone networks, the inference time difference before and after adding our RXFOOD is still trivial.

\begin{table}[ht]
    \caption{Average inference time for one 256$\times$256$\times$3 color image on the RGB-N task.}
	\label{tab:time}
	\centering
	\resizebox{0.5\columnwidth}{!}
	{
	\begin{tabular}{l|c}
		\toprule
		  Method & Time (s)  \\
		\midrule
		TBNet$_\mathrm{BAS}$~\cite{song2020deep} & 0.035 \\
            TBNet$_\mathrm{BAS}$+RXFOOD & 0.037 \\
            \midrule
            TBNet$_\mathrm{CPD}$~\cite{song2020deep} & 0.023  \\
            TBNet$_\mathrm{CPD}$+AVG & 0.024 \\
            TBNet$_\mathrm{CPD}$+MAX & 0.024 \\
            TBNet$_\mathrm{CPD}$+S$^2$MA~\cite{liu2020S2MA} & 0.025  \\
            TBNet$_\mathrm{CPD}$+RXFOOD & 0.027 \\
		\bottomrule
	\end{tabular}
	}
\end{table}

\section{Conclusion}
In this paper, we proposed a unified feature fusion method RXFOOD for features across different scales within the same modality and from different modality branches simultaneously, while existing works only handle the single-scale branch fusion or single-branch multiscale fusion. The spatial energy and channel energy inside each feature are important clues for finding object of interest in image, since they reflect the inter-relationship among different spatial positions and channels. So we design \textit{Energy Exchange Module} for information exchange by using spatial and channel energy matrices. Our RXFOOD can be easily integrated with any dual-branch encoder-decoder based networks as a plug-in module. Experimental results on three RGB-X object of interest detection tasks demonstrate that the proposed fusion method is quite effective for improving the performance of original backbone networks. 

\ifCLASSOPTIONcaptionsoff
  \newpage
\fi



%



{\small
\bibliographystyle{ieee_fullname}
\bibliography{egbib}
}

\end{document}